\documentclass[11pt,a4paper]{article}
\usepackage[hyperref]{acl2019}
\usepackage{times}
\usepackage{latexsym}
\usepackage{color}
\usepackage{linguex}
\usepackage{graphicx}
\usepackage{url}
\usepackage{listings}
\usepackage[utf8]{inputenc}
\usepackage{flushend}

\aclfinalcopy 

\newcommand{\nlu}[0]{\textsc{nlu}}
\newcommand{\sds}[0]{\textsc{sds}}

\newcommand{\sium}[0]{\textsc{sium}}

\newcommand{\iu}[0]{\textsc{iu}}
\newcommand{\add}[0]{\textsc{ADD}}
\newcommand{\revoke}[0]{\textsc{REVOKE}}
\newcommand{\asr}[0]{\textsc{asr}}

\title{Incrementalizing RASA's Open-Source\\ Natural Language Understanding Pipeline}

\author{Andrew Rafla \\
 Department of Computer Science\\
  Boise State University \\
  1910 University Dr. \\
  \texttt{andrewrafla@}\\ \texttt{u.boisestate.edu} \\\And
  Casey Kennington \\
 Department of Computer Science\\
  Boise State University \\
  1910 University Dr. \\
 \texttt{caseykennington@}\\ \texttt{boisestate.edu} \\}

\date{}

\begin{document}
\maketitle
\begin{abstract}
As spoken dialogue systems and chatbots are gaining more widespread adoption, commercial and open-sourced services for natural language understanding are emerging. In this paper, we explain how we altered the open-source RASA natural language understanding pipeline to process incrementally (i.e., word-by-word), following the incremental unit framework proposed by Schlangen and Skantze. To do so, we altered existing RASA components to process incrementally, and added an update-incremental intent recognition model as a component to RASA. Our evaluations on the Snips dataset show that our changes allow RASA to function as an effective incremental natural language understanding service.
\end{abstract}

\section{Introduction}

There is no shortage of services that are marketed as natural language understanding (\nlu) solutions for use in chatbots, digital personal assistants, or spoken dialogue systems (\sds). Recently, \newcite{Braun2017} systematically evaluated several such services, including Microsoft LUIS,\footnote{\url{https://www.luis.ai}} IBM Watson Conversation, API.ai,\footnote{\url{https://www.api.ai}} wit.ai,\footnote{\url{https://www.wit.ai}} Amazon Lex,\footnote{\url{https://aws.amazon.com/lex}} and RASA \cite{Bocklisch2017}.\footnote{\url{https://www.rasa.ai}} More recently, \newcite{Liu2019b} evaluated LUIS, Watson, RASA, and DialogFlow using some established benchmarks.\footnote{\url{https://dialogflow.com/}} Some \nlu\ services work better than others in certain tasks and domains with a perhaps surprising pattern: RASA, the only fully open-source \nlu\ service among those evaluated, consistently performs on par with the commercial services. 

Though these services yield state-of-the-art performance on a handful of \nlu\ tasks, one drawback to \sds\ and robotics researchers is the fact that all of these \nlu\ solutions process input at the utterance level; none of them process \emph{incrementally} at the word-level. Yet, research has shown that humans comprehend utterances as they unfold \cite{Tanenhaus1995}. Moreover, when a listener feels they are missing some crucial information mid-utterance, they can interject with a clarification request, so as to ensure they and the speaker are maintaining common ground \cite{Clark1996}. Users who interact with \sds s perceive incremental systems as being more natural than traditional, turn-based systems \citep{Aist2006,Skantze2009,Asri2014}, offer a more human-like experience  \citep{Edlund2008b} and are more satisfying to interact with than non-incremental systems \citep{Aistetal:incrunder-short}. Users even prefer interacting with an incremental \sds\ when the system is less accurate or requires filled pauses while replying \cite{Baumann2013a} or operates in a limited domain as long as there is incremental feedback \cite{kennington-schlangen2016}.

In this paper, we report our recent efforts in making the RASA \nlu\ pipeline process incrementally. We explain briefly the RASA framework and pipeline, explain how we altered the RASA framework and individual components (including a new component which we added) to allow it to process incrementally, then we explain how we evaluated the system to ensure that RASA works as intended and how researchers can leverage this tool.

\section{The RASA NLU Pipeline}

RASA consists of \nlu\ and \emph{core} modules, the latter of which is akin to a dialogue manager; our focus here is on the \nlu. The \nlu\ itself is further modularized as \emph{pipelines} which define how user utterances are processed, for example an utterance can pass through a tokenizer, named entity recognizer, then an intent classifier before producing a distribution over possible dialogue acts or intents. The pipeline and the training data are authorable (following a markdown representation; json format can also be used for the training data) allowing users to easily setup and run experiments in any domain as a standalone \nlu\ component or as a module in a \sds\ or chatbot. Importantly, RASA has provisions for authoring new components as well as altering existing ones. 

\begin{figure}[ht!]
    \includegraphics[width=0.48\textwidth,keepaspectratio=true,height=\textheight]{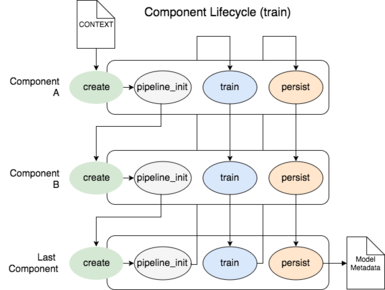}
    \caption[RASA component lifecycle]{The lifecycle of RASA components (from \url{https://rasa.com/docs/nlu/})}
    \label{figure:rasa-component}
\end{figure}

Figure~\ref{figure:rasa-component} shows a schematic of a pipeline for three components. The context (i.e., training data) is passed to Component A which performs its training, then persists a trained model for that component. Then the data is passed through Component A as input for Component B which also trains and persists, and so on for Component C. During runtime, the persisted models are loaded into memory and together form the \nlu\ module.

\section{Incrementalizing RASA}

Our approach to making RASA incremental follows the \emph{incremental unit} (\iu) framework \newcite{Schlangen2011} as has been done in previous work for dialogue processing toolkits \cite{Baumann2012}. We treat each module in RASA as an \iu\ processing module and specifically make use of the \add\ and \revoke\ \iu\ operations; for example, \add\ when a new word is typed or recognized by a speech recognizer, and \revoke\ if that word is identified as having been erroneously recognized in light of new information. 

By default, RASA components expect full utterances, not single words. In addition to the challenge of making components in the \nlu\ pipeline process word-by-word, we encounter another important problem: there is no ready-made signal for the end of an utterance. To solve this, we added functionality to signal the end of an utterance; this signal can be triggered by any component, including the speech recognizer where it has traditionally originated via endpointing. With this flexibility, any component (or set of components) can make a more informed decision about when an utterance is complete (e.g., if a user is uttering installments, endpointing may occur, but the intent behind the user's installments is not yet complete; the decision as to when an utterance is complete can be made by the \nlu\ or dialogue manager). 

Training RASA \nlu\ proceeds as explained above (i.e., non-incrementally). For runtime, processing incrementally through the RASA pipeline is challenging because each component must have provisions for handling word-level input and must be able to handle \add\ and \revoke\ \iu\ operations. Each component in a pipeline, for example, as depicted in Figure~\ref{figure:rasa-component}, must operate in lock-step with each other where a word is \add ed to Component A which beings processing immediately, then \add s its processing result to Component B, then Component B processes and passes output to Component C all before the next word is produced for Component A.


\subsection{Incrementalizing RASA Components}

We now explain how we altered specific RASA components to make them work incrementally.

\paragraph{Message} The \texttt{Message} class in RASA \nlu\ is the main message bus between components in the pipeline. \texttt{Message} follows a blackboard approach to passing information between components. For example, in a pipeline containing a tokenizer, intent classifier, and entity extractor, each of the components would store the tokens, intent class, and entities in the \texttt{Message} object, respectively. Our modifications to \texttt{Message} were minimal; we simply used it to store \iu s and corresponding edit types (i.e., \add\ or \revoke). 

\paragraph{Component} In order to incrementalize RASA \nlu, we extended the base \texttt{Component} to make an addition of a new component, \texttt{IncrementalComponent}. A user who defines their own \texttt{IncrementalComponent} understands the difference in functionality, notably in the \texttt{parse} method. At runtime, a non-incremental component expects a full utterance, whereas an incremental one expects only a single \iu. Because non-incremental components expect the entire utterance, they have no need to save any internal state across \texttt{process} calls, and can clear any internal data at the end of the method. However, with incremental components, that workflow changes; each call to \texttt{process} must maintain its internal state, so that it can be updated as it receives new \iu s. Moreover, \texttt{IncrementalComponent}s additionally have a \texttt{new\_utterance} method. In non-incremental systems, the call to \texttt{process} implicitly signals that the utterance has been completed, and there is no need to store internal data across \texttt{process} calls, whereas incremental systems lose that signal as a result. The \texttt{new\_utterance} method acts as that signal.

\paragraph{Interpreter} The \texttt{Interpreter} class in RASA \nlu\ is the main interface between user input (e.g., \asr) and the series of components in the pipeline. On training, the \texttt{Interpreter} prepares the training data, and serially calls train on each of the components in the pipeline. Similarly, to process input, one uses the \texttt{Interpreter}’s parse method, where the \texttt{Interpreter} prepares the input (i.e., the ongoing utterance) and serially calls process on the components in the pipeline (analgous to left buffer updates in the \iu\ framework). As a result of its design, we were able to leverage the \texttt{Interpreter} class for incremental processing, notably because of its use of a persistent \texttt{Message} object as a bus of communication between Components.

As with our implementation of the \texttt{IncrementalComponent} class, we created the \texttt{IncrementalInterpreter}. The \texttt{IncrementalInterpreter} class adds two new methods:

\begin{itemize}
    \item \texttt{new\_utterance}
    \item \texttt{parse\_incremental}
\end{itemize}

The \texttt{new\_utterance} method is fairly straightforward; it clears RASA \nlu’s internal \texttt{Message} object that is shared between components, and calls each \texttt{IncrementalComponent} in the pipeline’s \texttt{new\_utterance} method, signaling that the utterance has been completed, and for each component to clear their internal states. The \texttt{parse\_incremental} method takes the \iu\ from the calling input (e.g., \asr), and appends it to a list of previous \iu s being stored in the \texttt{Message} object. After the \iu\ has been added to the Message, the \texttt{IncrementalInterpreter} calls each component’s \texttt{process} method, where they can operate on the newest \iu. This was intentionally designed to be generalizable, so that future incremental components can use different formats or edit types for their respective \iu\ framework implementation. 


\begin{table*}[!ht]
\centering
\begin{tabular}{|l|l|l|}
\hline
\textbf{Implementation}      & \textbf{Intent F1-Score} & \textbf{Entities F1-Score} \\ \hline
Tensorflow (non-incremental) & 0.93                     & 0.86                       \\ \hline
Tensorflow (restart-incremental) & 0.93                 & 0.85                       \\ \hline
SIUM (non-incremental)       & 0.37                     & 0.34                       \\ \hline
SIUM (update-incremental)           & 0.36                     & 0.34                       \\ \hline
\end{tabular}
\caption{Test data results of non-incremental TensorFlow, restart-incremental TensorFlow, non-incremental SIUM, and update-incremental SIUM. \label{tbl:results}}
\end{table*}

\subsection{Incremental Intent Recognizer Components}

\paragraph{Simple Incremental Update Model} With the incremental framework in place, we further developed a sample incremental component to test the functionality of our changes. For this, we used the Simple Incremental Update Model (\sium) described in \cite{Kennington2017a}. This model is a generative factored joint distribution, which uses a simple Bayesian update as new words are added. At each \iu, a distribution of intents and entities are generated with confidence scores, and the intent can be classified at each step as the output with the highest confidence value. Entities on the other hand, can be extracted if their confidence exceeds a predetermined threshold. 


\paragraph{Restart-Incremental TensorFlow Embeddings}  Following \newcite{khouzaimi-laroche-lefevre:2014:W14-43}), we incrementalizaed RASA's existing Tensorflow Embedding component for intent recognition as an incremental component. The pipeline consists of a whitespace tokenizer, scikit-learn Conditional Random Field (\textsc{crf}) entity extractor, Bag-of-Words featurizer, and lastly, a TensorFlow Neural Network for intent classification. To start with incrementalizing, we modified the whitespace tokenizer to work on word-level increments, rather than the entire utterance. For the \textsc{crf} entity extractor, we modified it to update the entities up to that point in the utterance with each process call, and then modified the Bag-of-Words featurizer to update its embeddings with each process call by vectorizing the individual word in the \iu, and summing that vector with the existing embeddings. At each word \iu\ increment, we treat the entire utterance prefix to that point as a full utterance as input to the Tensorflow Embeddings component, which returns a distribution over intents. This process is repeated until all words in the utterance have been added to the prefix. In this way, the component differs from \sium\ in that it doesn't update its internal state; rather, it treats each prefix as a full utterance (i.e., so-called \emph{restart}-incrementality). 

\section{Experiment}

In this section, we explain a simple experiment we conducted to evaluate our work in incrementalizing RASA by using the update-incremental \sium\ and restart-incremental tensorflow-embedding modules in a known \nlu\ task. 

\subsection{Data, Task, Metrics}

To evaluate the performance of our approach, we used a subset of the SNIPS \cite{Coucke2018} dataset, which is readily available in RASA \nlu\ format. Our training data consisted of 700 utterances, across 7 different intents (\texttt{AddToPlaylist}, \texttt{BookRestaurant}, \texttt{GetWeather}, \texttt{PlayMusic}, \texttt{RateBook}, \texttt{SearchCreativeWork}, and \linebreak \texttt{SearchScreeningEvent}). In order to test our implementation of incremental components, we initially benchmarked their non-incremental counterparts, and used that as a baseline for the incremental versions (to treat the \sium\ component as non-incremental, we simply applied all words in each utterance to it and obtained the distribution over intents after each full utterance had been processed). 

We use accuracy of intent and entity recognition as our task and metric. To evaluate the components worked as intended, we then used the \texttt{IncrementalInterpreter} to parse the messages as individual \iu s. To ensure \revoke\ worked as intended, we injected random incorrect words at a rate of 40\%, followed by subsequent revokes, ensuring that an \add\ followed by a revoke resulted in the same output as if the incorrect word had never been added. While we implemented both an update-incremental and a restart-incremental RASA \nlu\ component, the results of the two cannot be directly compared for accuracy as the underlying models differ greatly (i.e., \sium\ is generative, whereas Tensorflow Embedding is a discriminative neural network; moreover, \sium\ was designed to work as a reference resolution component to physical objects, not abstract intents), nor are these results conducive to an argument of update- vs. restart-incremental approaches, as the underlying architecture of the models vary greatly. 

\subsection{Results}

The results of our evaluation can be found in Table \ref{tbl:results}. These results show that our incremental implementation works as intended, as the incremental and non-incremental version of each component yieled the same results. While there is a small variation between the F1 scores between the non-incremental and incremental components, 1\% is well within a reasonable tolerance as there is some randomness in training the underlying model.

\section{Conclusion}

RASA \nlu\ is a useful and well-evaluated toolkit for developing \nlu\ components in \sds\ and chatbot systems. We extended RASA by adding provisions for incremental processing generally, and we implemented two components for intent recognition that used update- and restart-incremental approaches. Our results show that the incrementalizing worked as expected. For ongoing and future work, we plan on developing an update-incremental counterpart to the Tensorflow Embeddings component that uses a recurrent neural network to maintain the state. We will further evaluate our work with incremental \asr\ in live dialogue tasks. We will make our code available upon acceptance of this publication. 

\bibliography{refs}

\begin{thebibliography}{17}
\expandafter\ifx\csname natexlab\endcsname\relax\def\natexlab#1{#1}\fi

\bibitem[{Aist et~al.(2006)Aist, Allen, Campana, Galescu, Gallo, Stoness,
  Swift, and Tanenhaus}]{Aist2006}
Gregory Aist, James Allen, Ellen Campana, Lucian Galescu, Carlos Gallo, Scott
  Stoness, Mary Swift, and Michael Tanenhaus. 2006.
\newblock {Software architectures for incremental understanding of human
  speech}.
\newblock In \emph{Proceedings of CSLP}, pages 1922----1925.

\bibitem[{Aist et~al.(2007)Aist, Allen, Campana, Gallo, Stoness, and
  Swift}]{Aistetal:incrunder-short}
Gregory Aist, James Allen, Ellen Campana, Carlos~Gomez Gallo, Scott Stoness,
  and Mary Swift. 2007.
\newblock {Incremental understanding in human-computer dialogue and
  experimental evidence for advantages over nonincremental methods}.
\newblock In \emph{Pragmatics}, volume~1, pages 149--154, Trento, Italy.

\bibitem[{Asri et~al.(2014)Asri, Laroche, Pietquin, and Khouzaimi}]{Asri2014}
Layla~El Asri, Romain Laroche, Olivier Pietquin, and Hatim Khouzaimi. 2014.
\newblock {NASTIA: Negotiating Appointment Setting Interface}.
\newblock In \emph{Proceedings of LREC}, pages 266--271.

\bibitem[{Baumann(2013)}]{Baumann2013a}
Timo Baumann. 2013.
\newblock \emph{{Incremental spoken dialogue processing: Architecture and
  lower-level components}}.
\newblock Ph.D. thesis, Bielefeld University.

\bibitem[{Baumann and Schlangen(2012)}]{Baumann2012}
Timo Baumann and David Schlangen. 2012.
\newblock {The InproTK 2012 release}.
\newblock In \emph{NAACL-HLT Workshop on Future directions and needs in the
  Spoken Dialog Community: Tools and Data (SDCTD 2012)}, pages 29--32.

\bibitem[{Bocklisch et~al.(2017)Bocklisch, Ai, Faulkner, Ai, Pawlowski, Ai,
  Nichol, and Ai}]{Bocklisch2017}
Tom Bocklisch, Rasa~Tom@rasa Ai, Joey Faulkner, Rasa~Joey@rasa Ai, Nick
  Pawlowski, Rasa~Nick@rasa Ai, Alan Nichol, and Rasa~Alan@rasa Ai. 2017.
\newblock \href
  {http://alborz-geramifard.com/workshops/nips17-Conversational-AI/Papers/17nipsw-cai-rasa.pdf}
  {{Rasa: Open Source Language Understanding and Dialogue Management}}.
\newblock In \emph{Proceedings of the 31st Conference on Neural Information
  Processing Systems}, Long Beach, CA.

\bibitem[{Braun et~al.(2017)Braun, {Hernandez Mendez}, Matthes, and
  Langen}]{Braun2017}
Daniel Braun, Adrian {Hernandez Mendez}, Florian Matthes, and Manfred Langen.
  2017.
\newblock \href {https://www.wit.ai} {{Evaluating Natural Language
  Understanding Services for Conversational Question Answering Systems}}.
\newblock In \emph{Proceedings of the SIGDIAL 2017 Conference}, pages 174--185.

\bibitem[{Clark(1996)}]{Clark1996}
Herbert~H Clark. 1996.
\newblock \href {https://doi.org/10.2277/0521561582} {\emph{{Using Language}}}.
\newblock Cambridge University Press.

\bibitem[{Coucke et~al.(2018)Coucke, Saade, Ball, Bluche, Caulier, Leroy,
  Doumouro, Gisselbrecht, Caltagirone, Lavril, Primet, and Dureau}]{Coucke2018}
Alice Coucke, Alaa Saade, Adrien Ball, Th{\'{e}}odore Bluche, Alexandre
  Caulier, David Leroy, Cl{\'{e}}ment Doumouro, Thibault Gisselbrecht,
  Francesco Caltagirone, Thibaut Lavril, Ma{\"{e}}l Primet, and Joseph Dureau.
  2018.
\newblock \href {http://arxiv.org/abs/1805.10190} {{Snips Voice Platform: an
  embedded Spoken Language Understanding system for private-by-design voice
  interfaces}}.
\newblock \emph{arXiv}.

\bibitem[{Edlund et~al.(2008)Edlund, Gustafson, Heldner, and
  Hjalmarsson}]{Edlund2008b}
Jens Edlund, Joakim Gustafson, Mattias Heldner, and Anna Hjalmarsson. 2008.
\newblock \href {https://doi.org/10.1016/j.specom.2008.04.002} {{Towards
  human-like spoken dialogue systems}}.
\newblock \emph{Speech Communication}, 50(8-9):630--645.

\bibitem[{Kennington and Schlangen(2017)}]{Kennington2017a}
C.~Kennington and D.~Schlangen. 2017.
\newblock \href {https://doi.org/10.1016/j.csl.2016.04.002} {{A simple
  generative model of incremental reference resolution for situated dialogue}}.
\newblock \emph{Computer Speech and Language}, 41.

\bibitem[{Kennington and Schlangen(2016)}]{kennington-schlangen2016}
Casey Kennington and David Schlangen. 2016.
\newblock \href {http://www.aclweb.org/anthology/W16-3631} {{Supporting Spoken
  Assistant Systems with a Graphical User Interface that Signals Incremental
  Understanding and Prediction State}}.
\newblock In \emph{Proceedings of the 17th Annual Meeting of the Special
  Interest Group on Discourse and Dialogue}, pages 242--251, Los Angeles.
  Association for Computational Linguistics.

\bibitem[{Khouzaimi et~al.(2014)Khouzaimi, Laroche, and
  Lefevre}]{khouzaimi-laroche-lefevre:2014:W14-43}
Hatim Khouzaimi, Romain Laroche, and Fabrice Lefevre. 2014.
\newblock \href {http://www.aclweb.org/anthology/W14-4314} {{An easy method to
  make dialogue systems incremental}}.
\newblock In \emph{Proceedings of the 15th Annual Meeting of the Special
  Interest Group on Discourse and Dialogue (SIGDIAL)}, June, pages 98--107,
  Philadelphia, PA, U.S.A. Association for Computational Linguistics.

\bibitem[{Liu et~al.(2019)Liu, Eshghi, Swietojanski, and Rieser}]{Liu2019b}
Xingkun Liu, Arash Eshghi, Pawel Swietojanski, and Verena Rieser. 2019.
\newblock \href {http://arxiv.org/abs/1903.05566} {{Benchmarking Natural
  Language Understanding Services for building Conversational Agents}}.

\bibitem[{Schlangen and Skantze(2011)}]{Schlangen2011}
David Schlangen and Gabriel Skantze. 2011.
\newblock \href {https://doi.org/10.5087/dad.2011.105} {{A General, Abstract
  Model of Incremental Dialogue Processing}}.
\newblock In \emph{Dialogue {\&} Discourse}, volume~2, pages 83--111.

\bibitem[{Skantze and Schlangen(2009)}]{Skantze2009}
Gabriel Skantze and David Schlangen. 2009.
\newblock \href {https://doi.org/10.3115/1609067.1609150} {{Incremental
  dialogue processing in a micro-domain}}.
\newblock \emph{Proceedings of the 12th Conference of the European Chapter of
  the Association for Computational Linguistics on EACL 09}, (April):745--753.

\bibitem[{Tanenhaus et~al.(1995)Tanenhaus, Spivey-Knowlton, Eberhard, and
  Sedivy}]{Tanenhaus1995}
Michael Tanenhaus, Michael Spivey-Knowlton, Kathleen Eberhard, and Julie
  Sedivy. 1995.
\newblock \href {https://doi.org/10.1126/science.7777863} {{Integration of
  visual and linguistic information in spoken language comprehension.}}
\newblock \emph{Science (New York, N.Y.)}, 268(5217):1632--1634.

\end{thebibliography}
\bibliographystyle{acl_natbib}

\appendix

\section{Appendix}
\label{sec:appendixA}

{\small
\begin{verbatim}
language: "en"
pipeline:
- name: "intent_featurizer_count_vectors"
- name: "intent_..._tensorflow_embedding"
  intent_tokenization_flag: true
  intent_split_symbol: "+"
\end{verbatim}
}

\end{document}